\documentclass{article}

\usepackage[preprint]{style_2026}

\usepackage[utf8]{inputenc}
\usepackage[T1]{fontenc}
\usepackage{hyperref}
\usepackage{url}
\usepackage{booktabs}
\usepackage{amsfonts}
\usepackage{nicefrac}
\usepackage{microtype}
\usepackage{xcolor}
\usepackage{graphicx}
\usepackage{multirow}
\usepackage{algorithm}
\usepackage{algorithmic}
\usepackage{amsmath}
\usepackage{wrapfig}
\usepackage{caption}
\usepackage{wrapfig}
\usepackage{float}
\usepackage{tabularx}
\usepackage{url}
\usepackage{hyperref}
\usepackage{natbib}

\title{Topology-Preserving Polygon Augmentation for Segmentation in Structured Visual Domains}

\author{%
  Sudip Laudari\thanks{Corresponding author.} \\
  Independent Researcher \\
  \texttt{sudip.laudari@sydney.edu.au} \\
  \And
  Sang Hun Baek \\
  Independent Researcher \\
  \texttt{shawnbaek@kookmin.ac.kr} \\
}

\begin{document}

\maketitle


\begin{abstract}
Geometric data augmentation is widely used in segmentation workflows, but polygon annotations are often assumed to remain valid after transformation. This assumption can fail in structured domains such as architectural floorplan analysis, where a region may contain an interior void encoded as part of a single ordered polygon chain. Cropping or clipping can remove bridge vertices in this chain, causing one semantic region to split into disconnected components. We propose a lightweight topology-preserving augmentation strategy that repairs missing adjacency relations in index space while preserving the original vertex order. The method adds minimal overhead and can be integrated into existing preprocessing workflows. Experiments show that the proposed approach achieves near-perfect Cyclic Adjacency Preservation (CAP) across common geometric transformations and improves annotation consistency in polygon-based segmentation.
\end{abstract}



\section{Introduction}
Image segmentation is fundamental to extracting spatial structures and semantic boundaries from complex scenes \citep{long2015fully, ronneberger2015u}. Although many standard benchmarks focus on simply connected objects, real-world structured domains such as architectural layouts, satellite imagery, and medical scans often contain regions with interior voids \citep{liu2018floornet, dodge2017parsing}. In these settings, polygon-based annotations are widely used because they provide compact boundary representations and support downstream tasks such as vectorization and geometric reconstruction \citep{lin2014microsoft, liu2018floornet, acuna2018efficient, dodge2017parsing, girard2021polygonal}.

\begin{wrapfigure}{r}{0.48\textwidth}
\vspace{-8pt}
\centering
\includegraphics[width=0.46\textwidth]{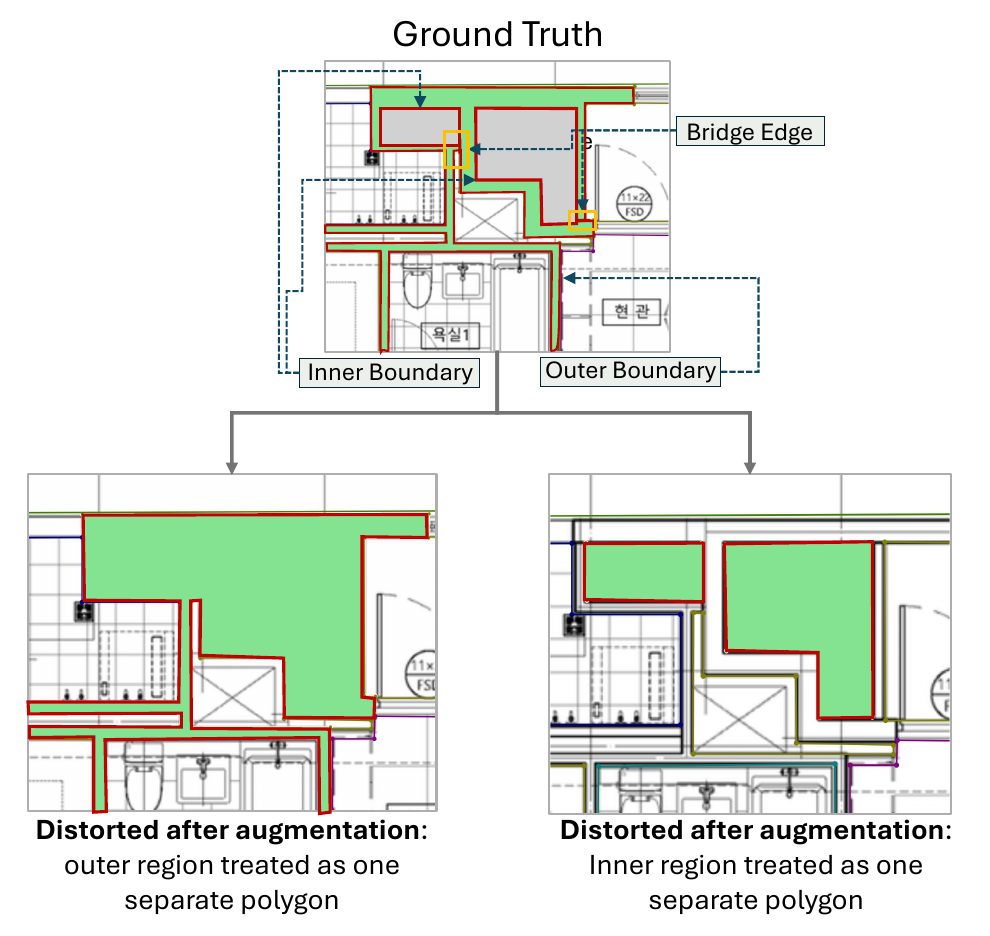}
\caption{Failure of ring-type polygon annotations after geometric augmentation. The ground-truth region is encoded as a single cyclic polygon chain in which bridge edges connect the outer and inner boundaries. After augmentation, these bridge connections can be disrupted, causing the outer or inner region to be treated as a separate polygon.}
\label{fig:figure1}
\vspace{-10pt}
\end{wrapfigure}

To improve model generalization, these annotations are commonly transformed through geometric operations such as rotation, scaling, cropping, and translation \citep{buslaev2020albumentations}. Most existing tools focus on updating vertex coordinates and implicitly assume that the resulting polygon remains structurally valid. This assumption is problematic for ring-type annotations, where an outer boundary and an inner void may be encoded as a single cyclic vertex chain connected by bridge edges. As shown in Figure~\ref{fig:figure1}, clipping or cropping near these bridges can remove key vertices and break the encoded connectivity. Once this occurs, a single semantic object may be incorrectly split into disconnected components, producing inconsistent supervision for segmentation models \citep{cordts2016cityscapes, benenson2019large}.

A natural alternative is to rasterize polygons into masks before applying augmentation. While this avoids direct manipulation of polygon vertices, it introduces a different limitation: the transformed annotation becomes resolution-dependent. Narrow gaps or thin bridge structures may merge during rasterization, and converting the transformed mask back into a polygon can alter the original vertex ordering \citep{liang2020polytransform, lazarow2022instance}. Topology-aware loss functions address related issues at the prediction level by encouraging topological agreement between predicted masks and ground truth, for example through Betti-number-based objectives \citep{hu2019topology}. However, they do not prevent ground-truth polygon annotations from being corrupted during preprocessing.

Motivated by this limitation, we introduce a topology-preserving repair mechanism for polygon augmentation. The method applies geometric transformations in mask space while preserving the original vertex order through index-space repair. This allows the augmented annotation to remain connected even when some vertices are removed by clipping or cropping. Our contributions are three-fold: (i) we propose a lightweight repair module that can be integrated into existing augmentation workflows; (ii) we introduce Cyclic Adjacency Preservation (CAP), a metric for measuring annotation-level connectivity; and (iii) we show that topology-consistent augmentation improves segmentation performance.


\section{Related Work}

Data augmentation is a standard strategy for improving geometric invariance in deep learning models \citep{shorten2019survey, perez2017effectiveness, krizhevsky2012imagenet}. Libraries such as Albumentations \citep{buslaev2020albumentations}, imgaug \citep{jung2018imgaug}, Augmentor \citep{bloice2017augmentor}, and AugmenTory \citep{ghahremani2024augmentory} provide practical tools for transforming images and labels. These methods are widely used in detection and segmentation workflows, including the YOLO family \citep{bochkovskiy2020yolov4, myshkovskyi2025exponential} and automatic augmentation strategies \citep{zoph2020learning, chen2021scale}. Their main goal is to keep labels spatially aligned with images. However, for polygon annotations, spatial alignment alone does not guarantee that the original vertex connectivity is preserved.

Polygon annotations are commonly represented as ordered vertex sequences in dataset formats and annotation tools. COCO \citep{lin2014microsoft}, for example, supports polygon-based segmentation masks, while tools such as LabelMe \citep{russell2008labelme} allow users to manually trace object boundaries. This representation is compact and editable, but it can be fragile when vertex order carries structural meaning. Such cases arise in architectural parsing, map vectorization, and technical drawing analysis, where polygon chains may encode relationships between outer and inner boundaries \citep{dodge2017parsing, kaleva2019cubicasa5k, zheng2020structured3d}.

Topology-aware learning has also been studied in image segmentation. Persistent homology and topological data analysis provide tools for describing connected components, holes, and higher-order shape properties \citep{edelsbrunner2008persistent, carlsson2009topology}. Segmentation losses based on Betti numbers have been proposed to encourage agreement between predicted and ground-truth topology \citep{hu2019topology}. Other methods refine boundaries or recover polygonal outputs from masks, including polygon transformer and mask-supervised polygon regression approaches \citep{liang2020polytransform, lazarow2022instance}. These methods improve structure at the prediction stage, whereas our work focuses on preventing topology errors from entering the training data during augmentation.


\section{Method}

\begin{figure*}[t]
\centering
\includegraphics[width=\textwidth]{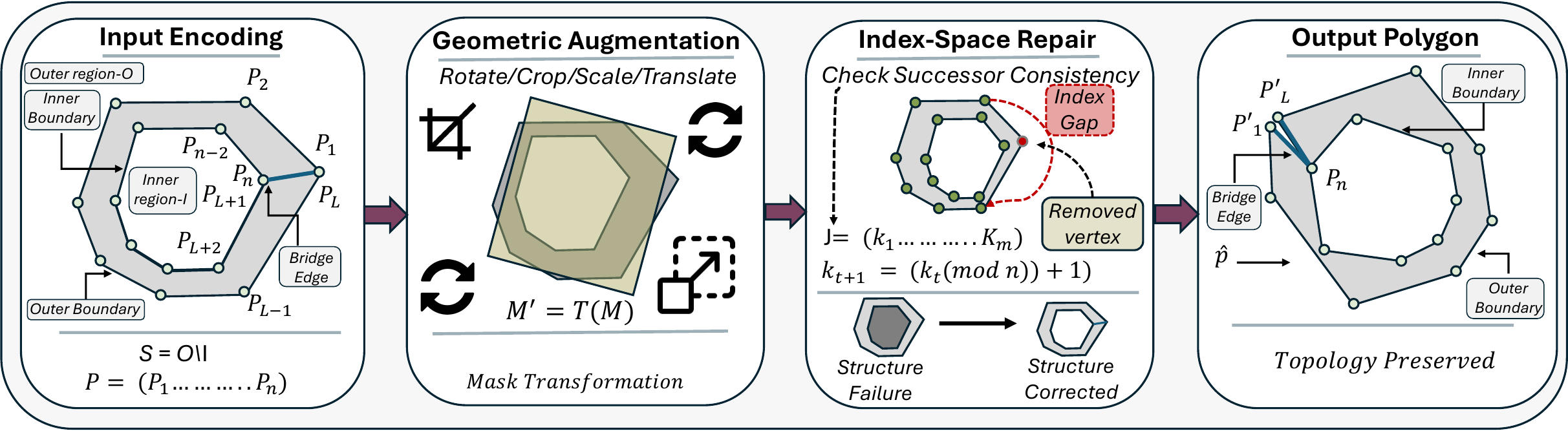}
\caption{Overview of the proposed topology-preserving augmentation framework. An input ring region is represented as a single ordered polygon chain $P=(p_1,\dots,p_n)$ spanning the outer and inner boundaries. The polygon is rasterized, transformed in mask space as $M'=T(M)$, and its surviving vertex indices are projected onto the augmented boundary. Missing index gaps are then repaired in index space, producing $\hat{P}$ while retaining the intended outer--inner boundary relationship.}
\label{fig:figure2}
\end{figure*}

Let $P = (p_1, p_2, \dots, p_n)$ denote a polygon annotation represented as a cyclic sequence of vertices, where $p_i \in \mathbb{R}^2$. We propose a four-stage topology-preserving augmentation framework: (1) mask rasterization, (2) mask-space transformation, (3) vertex index projection, and (4) connectivity repair. The overall procedure is shown in Figure~\ref{fig:figure2}, and the repair procedure is summarized in Algorithm~\ref{alg:ring_repair}.

\subsection{Ring-Type Polygon Encoding}

We consider a ring region defined as
\begin{equation}
S = O \setminus I ,
\label{eq:ring_region}
\end{equation}
where $O \subset \mathbb{R}^2$ is the outer region and $I \subset O$ is an interior hole. Although formats such as COCO often represent holes using multiple polygons, architectural annotations may encode both boundaries as a single cyclic chain. In this case, the polygon sequence is given by
\begin{equation}
P = (p_1, p_2, \dots, p_n),
\label{eq:polygon_sequence}
\end{equation}
where the vertices contain both the outer and inner boundaries of the ring region. We partition the sequence into
\begin{equation}
\mathcal{B}_{out} = (p_1,\dots,p_L), \qquad
\mathcal{B}_{in} = (p_{L+1},\dots,p_n).
\label{eq:boundary_partition}
\end{equation}
The two boundary segments are connected by a bridge edge
\begin{equation}
e_b = (p_L, p_{L+1}),
\label{eq:bridge_edge}
\end{equation}
and a closure edge
\begin{equation}
e_c = (p_n, p_1).
\label{eq:closure_edge}
\end{equation}
This encoding allows one ordered polygon to represent a region with an interior hole, but it also makes the annotation sensitive to vertex removal during augmentation.

\subsection{Mask-Based Geometric Augmentation}

\begin{wrapfigure}{r}{0.48\textwidth}
\vspace{-10pt}
\centering
\includegraphics[width=0.46\textwidth,keepaspectratio]{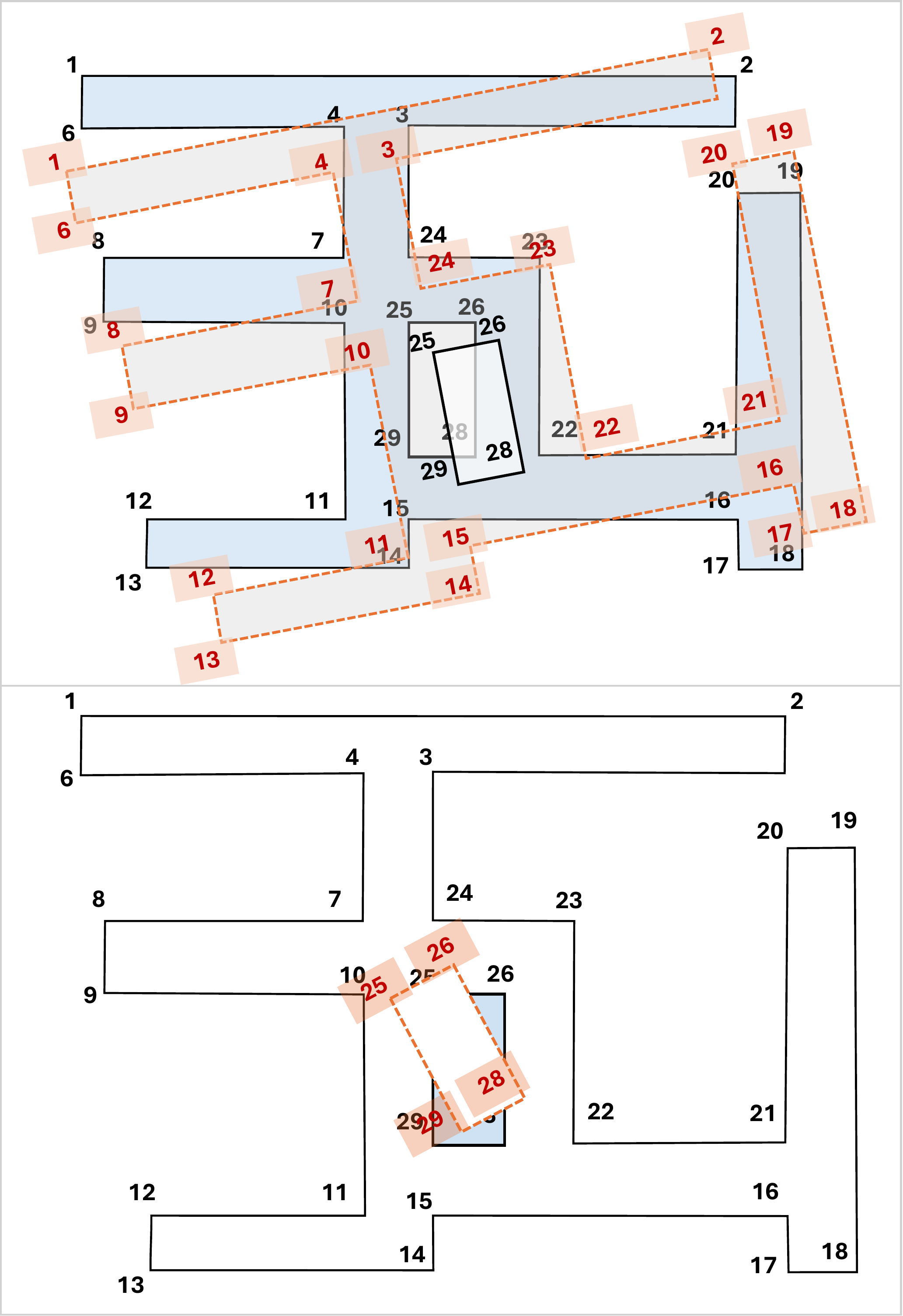}
\caption{Vertex projection after mask-based augmentation. The upper panel shows the transformed indexed polygon, where original vertices are mapped to the augmented boundary while retaining their indices. The lower panel shows the reconstructed polygon after invalid or clipped vertices are removed. This preserves index consistency during augmentation.}
\label{fig:projection}
\vspace{-70pt}
\end{wrapfigure}

Directly transforming polygon vertices can break connectivity when vertices are clipped by the image boundary. We therefore apply the geometric transformation in mask space, as illustrated in Figure~\ref{fig:figure2}. The polygon $P$ is first rasterized into a binary mask $M$, and the transformation is applied as
\begin{equation}
M' = T(M),
\label{eq:mask_transform}
\end{equation}
where $T$ denotes an affine operation such as rotation, scaling, translation, or cropping. Each original vertex keeps its index, which stores the traversal order of the polygon. After the mask-space transformation, surviving vertices are projected onto the boundary of $M'$. Vertices that fall outside the valid image or mask region are removed, while clipping intersections may be added as new boundary points when needed. The remaining vertices are then reconstructed according to their original index order, as illustrated in Figure~\ref{fig:projection}.

\subsection{Order-Preserving Connectivity Repair}

\begin{wrapfloat}{algorithm}{r}{0.52\textwidth}
\vspace{-8pt}
\caption{Order-Preserving Polygon Augmentation and Connectivity Repair}
\label{alg:ring_repair}
\vspace{2mm}
\small
\begin{algorithmic}[1]
\STATE \textbf{Input:} Original polygon $P = (p_1, p_2, \dots, p_n)$, augmentation transform $T$
\STATE \textbf{Output:} Repaired cyclic polygon chain $\hat{P}$

\STATE Rasterize $P$ into binary mask $M$
\STATE Apply geometric transformation: $M' \leftarrow T(M)$
\STATE Project original vertices onto $M'$ and extract surviving index set $J = (k_1, k_2, \dots, k_m)$
\STATE Sort $J$ according to the original cyclic order
\STATE Initialize $\hat{P} \leftarrow \emptyset$

\FOR{$t = 1$ to $m$}
    \STATE $i \leftarrow k_t$
    \STATE $j \leftarrow k_{((t \bmod m) + 1)}$
    \STATE Add directed edge $(p'_i, p'_j)$ to $\hat{P}$
\ENDFOR

\STATE \textbf{return} $\hat{P}$
\end{algorithmic}
\vspace{-1pt}
\end{wrapfloat}

Let $J = (k_1, k_2, \dots, k_m)$ denote the ordered indices of vertices that remain valid after augmentation, with $m \le n$ and $k_{m+1} := k_1$. We restore connectivity by linking consecutive surviving indices in their original cyclic order.

For each pair $(k_t, k_{t+1})$, we check whether the original successor relation is preserved:
\begin{equation}
k_{t+1} = (k_t \bmod n) + 1.
\label{eq:successor_relation}
\end{equation}
If this relation is violated, the missing indices correspond to vertices removed by clipping or cropping. Instead of treating this as a broken polygon, we reconnect the surviving vertices by adding the directed edge
\begin{equation}
(p'_{k_t}, p'_{k_{t+1}})
\label{eq:repair_edge}
\end{equation}
to the repaired polygon $\hat{P}$. This preserves the traversal direction while restoring connectivity among the remaining vertices.

The repair procedure in Algorithm~\ref{alg:ring_repair} implements this successor-based reconstruction. By sorting the surviving vertex indices according to their original cyclic order, the method avoids full contour reconstruction and preserves the index structure of the original polygon.

\subsection{Topological Consistency}
In the proposed representation, ring topology is encoded through successor relations in the vertex sequence. The bridge and closure edges define how the outer and inner boundary segments are connected within a single cyclic chain. Preserving adjacency among the surviving indices therefore maintains the structural transitions between $\mathcal{B}_{out}$ and $\mathcal{B}_{in}$. As a result, the repaired polygon $\hat{P}$ forms a single closed chain that remains consistent with the original ring-type annotation.

\subsection{Implementation and Complexity}

The repair procedure operates in $O(m)$ time, where $m$ is the number of surviving vertices after augmentation ($m \le n$). 
All operations are performed in index space and introduce negligible overhead compared with standard geometric transformations. 
Because the method avoids full contour reconstruction or polygon re-detection, it remains lightweight and can be integrated into existing augmentation workflows.

\section{Experimental Setup}

Our evaluation focuses on the structural integrity of polygon annotations after geometric augmentation. We compare whether standard workflows and the proposed repair method preserve the vertex connectivity of ring annotations when clipping or cropping removes vertices.

Unlike general segmentation studies that focus only on spatial overlap, such as mean Intersection over Union (mIoU), our experiments directly measure annotation-level topology. All methods use the same transformation ranges, and metrics are reported as class-wise averages across all samples.

\begin{wraptable}{r}{0.50\textwidth}
\vspace{-10pt}
\centering
\caption{Geometric augmentation parameters used in this study.}
\label{tab:augmentation_params}
\small
\begin{tabularx}{0.48\textwidth}{@{}l l X@{}}
\toprule
\textbf{Aug.} & \textbf{Parameter} & \textbf{Range} \\
\midrule
Rotation   & Angle       & $[-30^\circ, 30^\circ]$ \\
Scaling    & Scale       & $[0.7, 1.3]$ \\
Cropping   & Crop scale  & $[0.6, 1.0]$ \\
Rot. + Crop. & Angle / scale & $[-30^\circ, 30^\circ]$, $[0.6, 1.0]$ \\
Translation & Shift ratio & $[-0.1, 0.1]$ in $x,y$ \\
Flipping   & H / V flip  & Probability $= 0.5$ \\
\bottomrule
\end{tabularx}
\vspace{-30pt}
\end{wraptable}

\subsection{Augmentation Setup}

\paragraph{Augmentation Parameters:}
We evaluate common geometric transformations used in segmentation workflows, including rotation, scaling, translation, cropping, and horizontal flipping. The parameter ranges follow standard augmentation practice and are summarized in Table~\ref{tab:augmentation_params}. Color-based operations are excluded because they do not alter polygon geometry.

\paragraph{Baselines:}
We compare the proposed method with two commonly used augmentation workflows.

\begin{itemize}

\item \textbf{YOLOv11 Pipeline~\citep{yolov11}.}
This represents a typical training-time augmentation workflow in modern detection and segmentation frameworks. Images and polygon labels are transformed jointly, but cyclic vertex adjacency is not explicitly checked after clipping.

\item \textbf{Roboflow Augmentation~\citep{roboflow}.}
This represents a preprocessing-based dataset augmentation workflow. Polygon coordinates are transformed geometrically, but missing vertices are not repaired when clipping disrupts the original index order.

\item \textbf{Ours.}
Our method uses \textbf{Albumentations~\citep{buslaev2020albumentations}} for geometric transformations and applies the proposed index-space repair module to preserve cyclic connectivity.

\end{itemize}

\subsection{Evaluation Metric: Cyclic Adjacency Preservation (CAP)}

To quantify the structural integrity of polygon annotations, we introduce the \textbf{\emph{Cyclic Adjacency Preservation} (CAP)} metric. Let the original polygon be defined by the cyclic index set $I = \{1, \dots, n\}$ with successor function
\[
\mathrm{succ}(i) = (i \bmod n) + 1.
\]

After augmentation and boundary clipping, let

\[
J = (k_1, k_2, \dots, k_m)
\]
denote the ordered set of surviving vertex indices, where $m \le n$ and $k_{m+1} := k_1$. CAP measures the proportion of surviving vertex links that remain consistent with the original order:

\begin{equation}
\mathrm{CAP}(P \rightarrow P') =
\frac{1}{m}
\sum_{t=1}^{m}
\mathbb{I}[k_{t+1} = \mathrm{succ}(k_t)],
\end{equation}

where $\mathbb{I}[\cdot]$ denotes the indicator function.

The metric $\mathrm{CAP} \in [0,1]$ equals $1.0$ when all surviving vertices remain connected according to the original cyclic order. Importantly, CAP does not penalize valid vertex removal caused by cropping. Instead, it detects breaks in the remaining vertex sequence.

For a dataset containing $N$ ring-type polygon instances, we report the mean CAP:

\begin{equation}
\overline{\mathrm{CAP}} =
\frac{1}{N}
\sum_{i=1}^{N}
\mathrm{CAP}(P_i \rightarrow P'_i).
\end{equation}


\section{Results}

Table~\ref{tab:cap_results}(a) summarizes connectivity preservation under rotation. Baseline workflows show much lower CAP values, indicating that standard augmentation can disrupt vertex order when clipping removes intermediate points. In contrast, the proposed method achieves a CAP value close to $1.0$, showing that index-space repair keeps the ring structure intact after transformation.

\begin{wraptable}{r}{0.46\textwidth}
\vspace{-8pt}
\centering
\caption{CAP results for rotation and common geometric augmentations. Higher values indicate better preservation of cyclic polygon topology.}
\label{tab:cap_results}
\scriptsize

\textbf{(a) Rotation augmentation}\\[-0.3em]
\vspace{1mm}
\begin{tabular}{lc}
\toprule
\textbf{Method} & \textbf{CAP ($\uparrow$)} \\
\midrule
YOLO Aug. & 0.3278 \\
Roboflow  & 0.5497 \\
\textbf{Ours} & \textbf{0.9774} \\
\bottomrule
\end{tabular}

\vspace{0.6em}

\textbf{(b) Proposed repair across augmentations}\\[-0.3em]
\vspace{1mm}
\begin{tabular}{lc}
\toprule
\textbf{Augmentation} & \textbf{CAP ($\uparrow$)} \\
\midrule
Rotation             & 0.9774 \\
Cropping             & 0.9840 \\
Scaling              & 0.9827 \\
Flip                 & 0.9869 \\
Rotation + Cropping  & 0.9748 \\
\bottomrule
\end{tabular}

\vspace{-10pt}
\end{wraptable}

\begin{figure*}[!ht]
\centering
\includegraphics[width=0.95\linewidth]{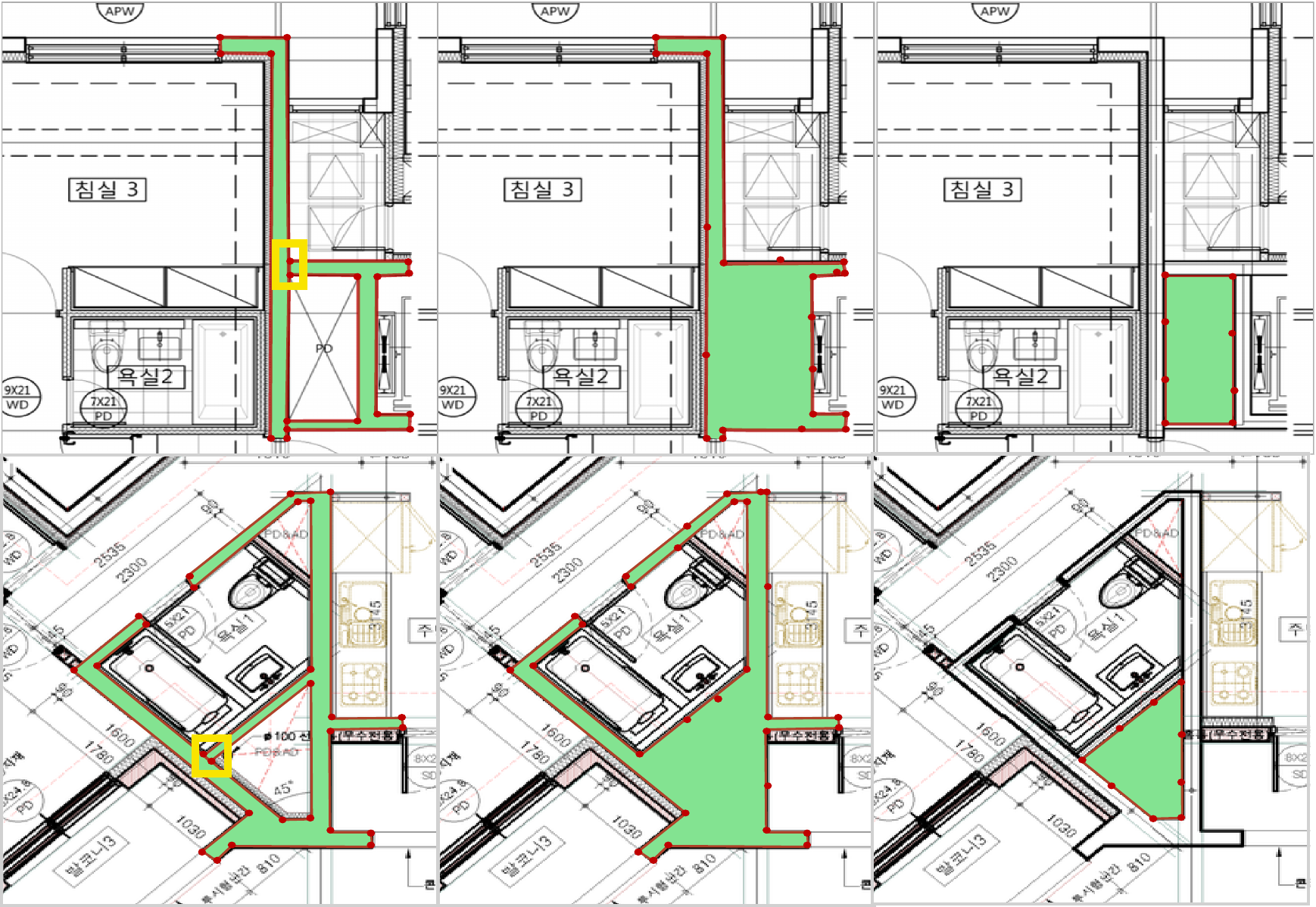}
\caption{Illustration of failure cases under Roboflow augmentation. The left column shows the ground-truth ring annotations, with bridge regions highlighted in yellow. The middle and right columns show augmented outputs in which the bridge connection between the outer and inner boundaries is lost, causing the original ring annotation to split into separate outer and inner polygon components.}
\label{label:figure3}
\end{figure*}

\begin{figure*}[!ht]
\centering
\includegraphics[width=0.95\linewidth]{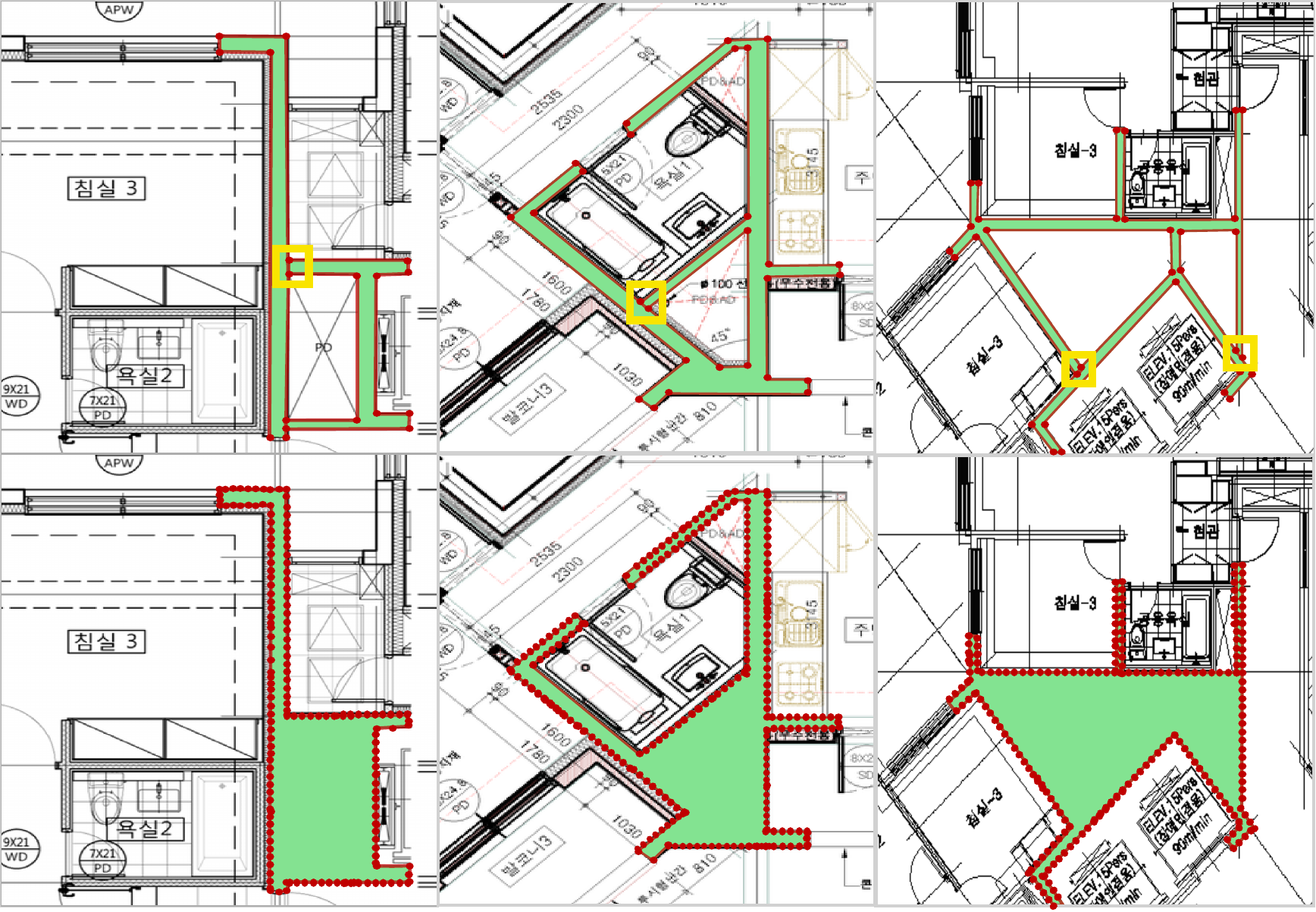}
\caption{Visualization of failure cases under YOLO training augmentation with default settings. The upper row shows the ground-truth ring annotations, with bridge regions highlighted in yellow, while the lower row shows the corresponding augmented outputs. After augmentation, the inner boundary is not preserved and the ring annotation is converted into a single outer polygon with dense boundary vertices. This changes the original sparse vertex ordering and removes the intended outer--inner boundary structure.}
\label{label:figure4}
\end{figure*}

\begin{figure*}[!ht]
\centering
\includegraphics[width=0.95\linewidth, height=0.55\linewidth]{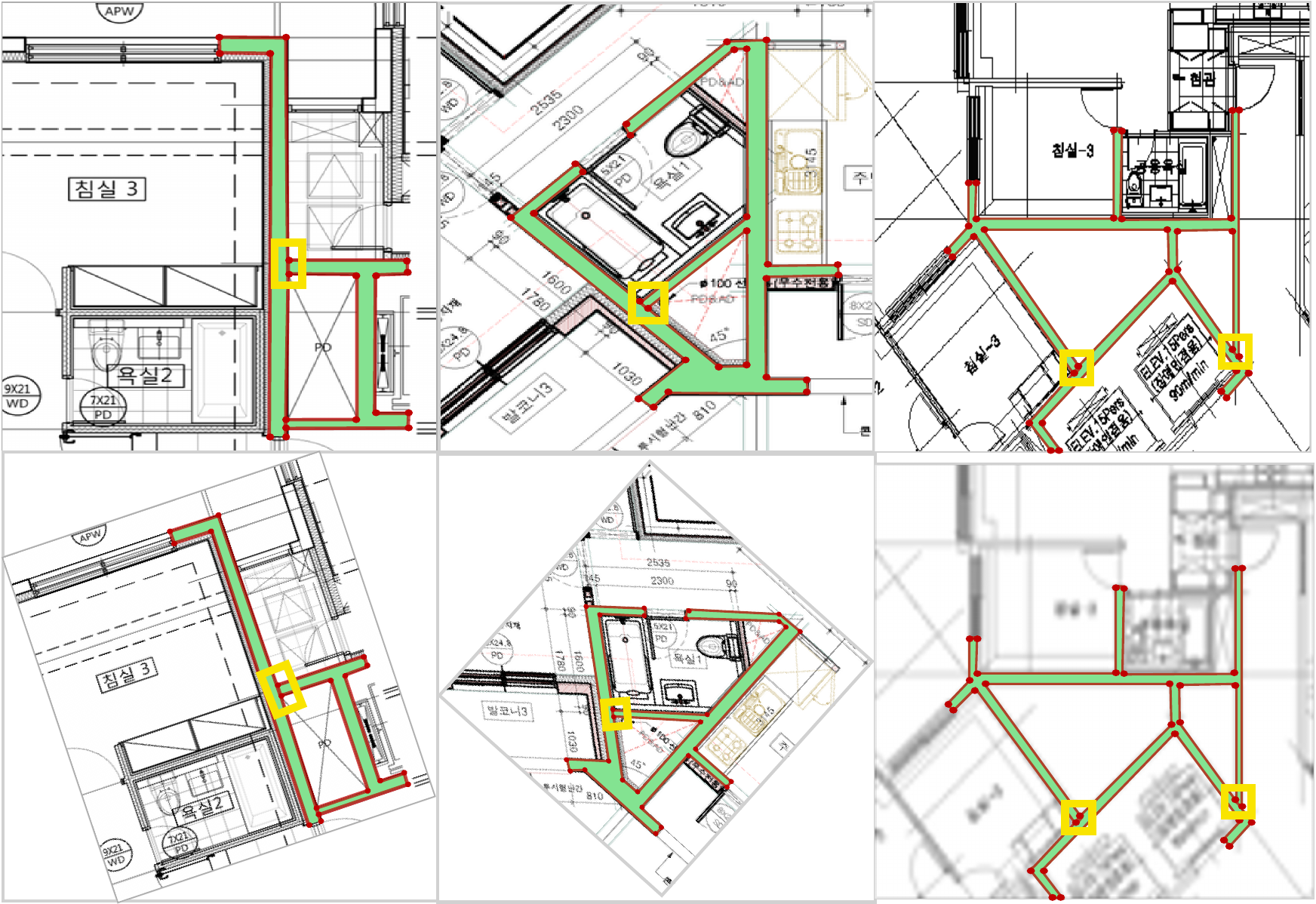}
\caption{Illustration of results produced by the proposed method. The upper row shows the ground-truth ring annotations, while the lower row shows the corresponding augmented outputs. The bridge regions remain preserved after augmentation, and the outer and inner boundaries continue to form a single connected ring structure.}
\label{label:figure5}
\end{figure*}

Qualitative comparisons further illustrate these differences (Figures~\ref{label:figure3}--\ref{label:figure5}). Roboflow augmentation can remove the bridge connection between the outer and inner boundaries, causing the original ring annotation to split into separate components (Figure~\ref{label:figure3}). YOLO augmentation shows a different failure mode: the output may contain many redundant contour vertices, changing the sparse structure of the original polygon (Figure~\ref{label:figure4}). This likely occurs when transformed masks are converted back into polygon contours rather than reconstructed from the original indexed vertices. In contrast, the proposed method preserves the original ordering and reconnects surviving vertices after augmentation (Figure~\ref{label:figure5}). The resulting annotations remain compact, connected, and consistent with the intended outer--inner boundary relationship.

Table~\ref{tab:cap_results}(b) reports performance across multiple transformations. CAP remains close to unity for all settings, including the more challenging rotation-and-cropping case. These results show that the proposed repair remains stable under common augmentation operations.

\paragraph{Effect of Topology Preservation on Segmentation Training:}

To assess the practical effect of cleaner annotations, we train YOLOv11-Seg and Mask R-CNN using datasets generated with standard augmentation and with the proposed topology-preserving method. All models are trained with identical hyperparameters and evaluated using mIoU.

Experiments are conducted on approximately 2000 Korean floorplan images with polygon-based annotations for architectural regions. The dataset is split into training, validation, and test sets using a 70\%, 20\%, and 10\% ratio. Augmentation is applied only to the training set. For each training image, five augmented samples are generated using the transformation ranges in Table~\ref{tab:augmentation_params}.

Table~\ref{tab:segmentation_results} reports the segmentation results.

\begin{table}[!h]
\centering
\caption{Segmentation performance under standard augmentation and the proposed topology-preserving augmentation. mIoU is reported in percentage points; higher is better.}
\label{tab:segmentation_results}
\small
\begin{tabular}{lcc}
\toprule
\textbf{Model} & \textbf{Augmentation} & \textbf{mIoU (\%) $\uparrow$} \\
\midrule
YOLOv11-Seg & Standard & 88.4 \\
YOLOv11-Seg & Ours & \textbf{90.8} \\
\midrule
Mask R-CNN & Standard & 85.2 \\
Mask R-CNN & Ours & \textbf{87.0} \\
\bottomrule
\end{tabular}
\end{table}

The dataset used in this study is publicly available through AI Hub\footnote{\url{https://aihub.or.kr/}}. Models trained with the proposed topology-preserving augmentation achieve higher mIoU for both architectures, suggesting that cleaner polygon annotations help segmentation models learn more reliable geometric boundaries.


\section{Code Availability}

The implementation is publicly available as a Python package and can be installed using:

\begin{verbatim}
pip install polyaug
\end{verbatim}

The source code and experimental scripts are available on GitHub\footnote{\url{https://github.com/Laudarisd/polyaug}}.


\section{Conclusion}

This work addressed a practical but often overlooked problem in segmentation with polygon annotations: geometric augmentation can silently damage annotation structure when clipping or cropping removes vertices from ordered polygon chains. We showed that this issue is especially important for ring annotations, where the relationship between outer and inner boundaries is encoded through vertex connectivity. To address this, we introduced a lightweight index-space repair method and the Cyclic Adjacency Preservation (CAP) metric for measuring annotation-level consistency. Experiments on floorplan segmentation show that the proposed method preserves connectivity across common transformations and improves downstream mIoU for both YOLOv11-Seg and Mask R-CNN. Future work will extend this idea to multi-ring objects, general vector annotations, and training pipelines that jointly preserve geometric structure during preprocessing and learning.



\bibliographystyle{plainnat}
\bibliography{ref}

\begin{thebibliography}{30}
\providecommand{\natexlab}[1]{#1}
\providecommand{\url}[1]{\texttt{#1}}
\expandafter\ifx\csname urlstyle\endcsname\relax
  \providecommand{\doi}[1]{doi: #1}\else
  \providecommand{\doi}{doi: \begingroup \urlstyle{rm}\Url}\fi

\bibitem[Acuna et~al.(2018)Acuna, Ling, Kar, and Fidler]{acuna2018efficient}
David Acuna, Huan Ling, Amlan Kar, and Sanja Fidler.
\newblock Efficient interactive annotation of segmentation datasets with {Polygon-RNN++}.
\newblock In \emph{Proceedings of the IEEE Conference on Computer Vision and Pattern Recognition (CVPR)}, pages 859--868, 2018.

\bibitem[Benenson et~al.(2019)Benenson, Popov, and Ferrari]{benenson2019large}
Rodrigo Benenson, Stefan Popov, and Vittorio Ferrari.
\newblock Large-scale interactive object segmentation with human annotators.
\newblock In \emph{Proceedings of the IEEE/CVF Conference on Computer Vision and Pattern Recognition (CVPR)}, pages 11700--11709, 2019.

\bibitem[Bloice et~al.(2017)Bloice, Stocker, and Holzinger]{bloice2017augmentor}
Marcus~D. Bloice, Christof Stocker, and Andreas Holzinger.
\newblock {Augmentor}: An image augmentation library for machine learning.
\newblock \emph{arXiv preprint arXiv:1708.04680}, 2017.

\bibitem[Bochkovskiy et~al.(2020)Bochkovskiy, Wang, and Liao]{bochkovskiy2020yolov4}
Alexey Bochkovskiy, Chien-Yao Wang, and Hong-Yuan~Mark Liao.
\newblock {YOLOv4}: Optimal speed and accuracy of object detection.
\newblock \emph{arXiv preprint arXiv:2004.10934}, 2020.

\bibitem[Buslaev et~al.(2020)Buslaev, Iglovikov, Khvedchenya, Parinov, Druzhinin, and Kalinin]{buslaev2020albumentations}
Alexander Buslaev, Vladimir~I. Iglovikov, Eugene Khvedchenya, Alex Parinov, Mikhail Druzhinin, and Alexandr~A. Kalinin.
\newblock {Albumentations}: Fast and flexible image augmentations.
\newblock \emph{Information}, 11\penalty0 (2):\penalty0 125, 2020.
\newblock \doi{10.3390/info11020125}.

\bibitem[Carlsson(2009)]{carlsson2009topology}
Gunnar Carlsson.
\newblock Topology and data.
\newblock \emph{Bulletin of the American Mathematical Society}, 46\penalty0 (2):\penalty0 255--308, 2009.
\newblock \doi{10.1090/S0273-0979-09-01249-X}.

\bibitem[Chen et~al.(2021)Chen, Li, Kong, Qi, Chu, Li, and Jia]{chen2021scale}
Yukang Chen, Yanwei Li, Tao Kong, Lu~Qi, Ruihang Chu, Lei Li, and Jiaya Jia.
\newblock Scale-aware automatic augmentation for object detection.
\newblock In \emph{Proceedings of the IEEE/CVF Conference on Computer Vision and Pattern Recognition (CVPR)}, pages 9563--9572, 2021.

\bibitem[Cordts et~al.(2016)Cordts, Omran, Ramos, Rehfeld, Enzweiler, Benenson, Franke, Roth, and Schiele]{cordts2016cityscapes}
Marius Cordts, Mohamed Omran, Sebastian Ramos, Timo Rehfeld, Markus Enzweiler, Rodrigo Benenson, Uwe Franke, Stefan Roth, and Bernt Schiele.
\newblock The {Cityscapes} dataset for semantic urban scene understanding.
\newblock In \emph{Proceedings of the IEEE Conference on Computer Vision and Pattern Recognition (CVPR)}, pages 3213--3223, 2016.

\bibitem[Dodge et~al.(2017)Dodge, Xu, and Stenger]{dodge2017parsing}
Samuel Dodge, Jiu Xu, and Bj{\"o}rn Stenger.
\newblock Parsing floor plan images.
\newblock In \emph{2017 Fifteenth IAPR International Conference on Machine Vision Applications (MVA)}, pages 358--361. IEEE, 2017.

\bibitem[Edelsbrunner et~al.(2008)Edelsbrunner, Harer, et~al.]{edelsbrunner2008persistent}
Herbert Edelsbrunner, John Harer, et~al.
\newblock Persistent homology: A survey.
\newblock \emph{Contemporary Mathematics}, 453\penalty0 (26):\penalty0 257--282, 2008.

\bibitem[Ghahremani et~al.(2024)Ghahremani, Hoseyni, Ahmadi, Mehrabi, and Nikoofard]{ghahremani2024augmentory}
Tanaz Ghahremani, Mohammad Hoseyni, Mohammad~Javad Ahmadi, Pouria Mehrabi, and Amirhossein Nikoofard.
\newblock {AugmenTory}: A fast and flexible polygon augmentation library.
\newblock \emph{arXiv preprint arXiv:2405.04442}, 2024.

\bibitem[Girard et~al.(2021)Girard, Smirnov, Solomon, and Tarabalka]{girard2021polygonal}
Nicolas Girard, Dmitriy Smirnov, Justin Solomon, and Yuliya Tarabalka.
\newblock Polygonal building extraction by frame field learning.
\newblock In \emph{Proceedings of the IEEE/CVF Conference on Computer Vision and Pattern Recognition (CVPR)}, pages 5891--5900, 2021.

\bibitem[Hu et~al.(2019)Hu, Li, Samaras, and Chen]{hu2019topology}
Xiaoling Hu, Fuxin Li, Dimitris Samaras, and Chao Chen.
\newblock Topology-preserving deep image segmentation.
\newblock \emph{Advances in Neural Information Processing Systems}, 32, 2019.

\bibitem[Jung(2018)]{jung2018imgaug}
Alexander Jung.
\newblock {imgaug}.
\newblock \url{https://github.com/aleju/imgaug}, 2018.

\bibitem[Kaleva and K{\"a}m{\"a}r{\"a}inen(2019)]{kaleva2019cubicasa5k}
Aleksi Kaleva and Joni-Kristian K{\"a}m{\"a}r{\"a}inen.
\newblock {CubiCasa5K}: A dataset and an improved multi-task model for floorplan image analysis.
\newblock In \emph{Scandinavian Conference on Image Analysis}, 2019.

\bibitem[Krizhevsky et~al.(2012)Krizhevsky, Sutskever, and Hinton]{krizhevsky2012imagenet}
Alex Krizhevsky, Ilya Sutskever, and Geoffrey~E. Hinton.
\newblock {ImageNet} classification with deep convolutional neural networks.
\newblock \emph{Advances in Neural Information Processing Systems}, 25, 2012.

\bibitem[Lazarow et~al.(2022)Lazarow, Xu, and Tu]{lazarow2022instance}
Justin Lazarow, Weijian Xu, and Zhuowen Tu.
\newblock Instance segmentation with mask-supervised polygonal regression transformers.
\newblock In \emph{Proceedings of the IEEE/CVF Conference on Computer Vision and Pattern Recognition (CVPR)}, 2022.

\bibitem[Liang et~al.(2020)Liang, Homayounfar, Ma, Xiong, Hu, and Urtasun]{liang2020polytransform}
Justin Liang, Namdar Homayounfar, Wei-Chiu Ma, Yuwen Xiong, Rui Hu, and Raquel Urtasun.
\newblock {PolyTransform}: Deep polygon transformer for instance segmentation.
\newblock In \emph{Proceedings of the IEEE/CVF Conference on Computer Vision and Pattern Recognition (CVPR)}, pages 9131--9140, 2020.

\bibitem[Lin et~al.(2014)Lin, Maire, Belongie, Hays, Perona, Ramanan, Doll{\'a}r, and Zitnick]{lin2014microsoft}
Tsung-Yi Lin, Michael Maire, Serge Belongie, James Hays, Pietro Perona, Deva Ramanan, Piotr Doll{\'a}r, and C.~Lawrence Zitnick.
\newblock Microsoft {COCO}: Common objects in context.
\newblock In \emph{European Conference on Computer Vision (ECCV)}, pages 740--755, 2014.
\newblock \doi{10.1007/978-3-319-10602-1_48}.

\bibitem[Liu et~al.(2018)Liu, Wu, and Furukawa]{liu2018floornet}
Chen Liu, Jiaye Wu, and Yasutaka Furukawa.
\newblock {FloorNet}: A unified framework for floorplan reconstruction from {3D} scans.
\newblock In \emph{Proceedings of the European Conference on Computer Vision (ECCV)}, pages 201--217, 2018.

\bibitem[Long et~al.(2015)Long, Shelhamer, and Darrell]{long2015fully}
Jonathan Long, Evan Shelhamer, and Trevor Darrell.
\newblock Fully convolutional networks for semantic segmentation.
\newblock In \emph{Proceedings of the IEEE Conference on Computer Vision and Pattern Recognition (CVPR)}, pages 3431--3440, 2015.

\bibitem[Myshkovskyi et~al.(2025)Myshkovskyi, Nazarkevych, Vysotska, and Yurynets]{myshkovskyi2025exponential}
Yurii Myshkovskyi, Mariia Nazarkevych, Victoria Vysotska, and Rostyslav Yurynets.
\newblock Exponential data augmentation methods for improving {YOLO} performance in computer vision tasks.
\newblock 2025.

\bibitem[Perez and Wang(2017)]{perez2017effectiveness}
Luis Perez and Jason Wang.
\newblock The effectiveness of data augmentation in image classification using deep learning.
\newblock \emph{arXiv preprint arXiv:1712.04621}, 2017.

\bibitem[{Roboflow, Inc.}(2023)]{roboflow}
{Roboflow, Inc.}
\newblock {Roboflow}: Annotation and augmentation platform.
\newblock \url{https://roboflow.com}, 2023.

\bibitem[Ronneberger et~al.(2015)Ronneberger, Fischer, and Brox]{ronneberger2015u}
Olaf Ronneberger, Philipp Fischer, and Thomas Brox.
\newblock {U-Net}: Convolutional networks for biomedical image segmentation.
\newblock In \emph{International Conference on Medical Image Computing and Computer-Assisted Intervention (MICCAI)}, pages 234--241. Springer, 2015.

\bibitem[Russell et~al.(2008)Russell, Torralba, Murphy, and Freeman]{russell2008labelme}
Bryan~C. Russell, Antonio Torralba, Kevin~P. Murphy, and William~T. Freeman.
\newblock {LabelMe}: A database and web-based tool for image annotation.
\newblock \emph{International Journal of Computer Vision}, 77\penalty0 (1):\penalty0 157--173, 2008.

\bibitem[Shorten and Khoshgoftaar(2019)]{shorten2019survey}
Connor Shorten and Taghi~M. Khoshgoftaar.
\newblock A survey on image data augmentation for deep learning.
\newblock \emph{Journal of Big Data}, 6\penalty0 (1):\penalty0 60, 2019.
\newblock \doi{10.1186/s40537-019-0197-0}.

\bibitem[{Ultralytics}(2024)]{yolov11}
{Ultralytics}.
\newblock {YOLOv11}: Real-time object detection.
\newblock \url{https://github.com/ultralytics/ultralytics}, 2024.

\bibitem[Zheng et~al.(2020)Zheng, Li, et~al.]{zheng2020structured3d}
Jia Zheng, Yangyan Li, et~al.
\newblock {Structured3D}: A large photo-realistic dataset for structured {3D} modeling.
\newblock In \emph{European Conference on Computer Vision (ECCV)}, 2020.

\bibitem[Zoph et~al.(2020)Zoph, Cubuk, Ghiasi, Lin, Shlens, and Le]{zoph2020learning}
Barret Zoph, Ekin~D. Cubuk, Golnaz Ghiasi, Tsung-Yi Lin, Jonathon Shlens, and Quoc~V. Le.
\newblock Learning data augmentation strategies for object detection.
\newblock In \emph{European Conference on Computer Vision (ECCV)}, pages 566--583. Springer, 2020.

\end{thebibliography}

\end{document}